\documentclass[conference]{IEEEtran}

\usepackage{cite}
\usepackage{amsmath,amssymb,amsfonts}
\usepackage{graphicx}
\usepackage{textcomp}
\usepackage{url}
\usepackage{balance}
\usepackage{booktabs}
\usepackage{array}
\usepackage{tabularx}
\usepackage{makecell}
\usepackage{tikz}
\usetikzlibrary{positioning,arrows.meta}

\newcommand{\twolinecaption}[2]{\caption{#1}}

\title{Where Quality Breaks in Compressed Short-Text Generation: Staged Bottleneck Localization}
\date{}
\author{
\IEEEauthorblockN{Alexey Gavrilov, Alan-Barsag Gazzaev, Sergey Muravyov}
\IEEEauthorblockA{ITMO University \\
Saint-Petersburg, Russia \\
alexgavrilov28@gmail.com, alanrbtx@gmail.com, mursmail@gmail.com}
}

\begin{document}
\maketitle

\begin{abstract}
Compressed short-text generators can fail in two different places: the codec may discard information before generation starts, or the latent generator may produce weak codes. Without separating these failure modes, researchers can spend compute improving the wrong component. We study this problem in a controlled 64-to-16 TinyStories case study built from a hierarchical VQ-VAE-2 codec and a masked discrete diffusion generator (MDLM). We use a staged validation protocol that separates codec reconstruction fidelity, latent generation quality, and auxiliary latent diagnostics under one shared external GPT-2 scorer, while reporting complementary semantic metrics for the geometry study. In the tested configuration, codec reconstruction alone raises median external perplexity from 15.17 to 27.36 (+80.4\%) and $p95$ from 25.10 to 98.91 (+294.1\%), showing that the dominant quality loss appears before latent generation begins. Under the same scorer, code-space MDLM remains materially stronger than token-space diffusion, reducing mean, median, and $p95$ by 32.9\%, 30.9\%, and 36.6\%, respectively. Geometry-aware regularization improves local latent proxies but does not improve decoded-text metrics in the available runs. The contribution is methodological rather than algorithmic: the paper presents a reusable staged diagnosis for one concrete pipeline and shows that, in this setting, codec fidelity rather than latent denoising sets the practical quality ceiling.
\end{abstract}

\section{Introduction}
A compressed text generator can fail before generation even starts. In a two-stage pipeline, quality may be lost because the codec discards information during compression or because the latent generator fails to produce a good code sequence. If these losses are not separated, it is easy to optimize the wrong component and still see no downstream gain.

This matters because compressed latent pipelines are attractive for short-text workloads: they shorten the modeled sequence, enable more parallel decoding, and can reduce the effective categorical burden relative to token-space generation. These practical advantages are only useful, however, if the main fidelity bottleneck is identified correctly.

We study this bottleneck-localization problem in one concrete system built from a hierarchical VQ-VAE-2 codec and a masked discrete diffusion generator (MDLM), asking which stage sets the practical fidelity ceiling under aggressive 64-to-16 compression. The main empirical picture is simple: code-space generation is materially better than token-space diffusion under the same external scorer, yet the dominant quality loss is already introduced by codec reconstruction.

The paper is intentionally positioned as a controlled methodological case study rather than a broad benchmark across datasets, architectures, or compression regimes. Our goal is not to propose a new training algorithm, but to make stage-wise failure attribution explicit. The main contributions are:
\begin{enumerate}
    \item We formulate a reusable staged validation protocol for compressed short-text generation that isolates representational loss from generation loss under one shared scorer.
    \item We show that in the tested 64-to-16 hierarchical discrete-latent pipeline, code-space generation is preferable to token-space diffusion, while the dominant practical bottleneck is codec reconstruction fidelity.
    \item We show that healthy codebook usage and stronger latent proxies are useful diagnostics, but they are not evidence of downstream text gain unless paired decoded-text metrics also improve.
\end{enumerate}

\section{Related Work}
Discrete latent modeling in text extends VQ-VAE and VQ-VAE-2 style representation learning, where a codec maps a long sequence into a shorter discrete representation \cite{vqvae,vqvae2}. In image generation, such tokenizers are widely used to make downstream generation cheaper. In text, the same factorization is attractive for short-sequence workloads, but semantic fidelity is more fragile because even a small reconstruction error can change meaning, syntax, or local factual detail.

For sequence generation in the latent domain, our implementation follows masked refinement and discrete diffusion ideas rather than fully autoregressive decoding. MDLM and related categorical diffusion or masked-refinement methods provide a practical way to denoise short discrete sequences in parallel \cite{mdlm,d3pm,sedd,ssdlm,diffusionlm,maskgit,maskpredict,argmaxflows}. Their engineering promise is straightforward: if the codec preserves enough information, then generation in a shorter code sequence can offer a better quality-efficiency trade-off than direct token-space diffusion.

Our evaluation setup combines one shared anchor metric with complementary semantic metrics. We use an external GPT-2 scorer because it can be applied consistently to originals, reconstructions, and generated text under one protocol \cite{gpt2}. For the geometry study, we additionally report SBERT similarity, BERTScore, MAUVE, and an LLM-judge summary as complementary text-level views rather than as interchangeable summary scores \cite{sbert,bertscore,mauve,llmjudge}. Our contribution is therefore methodological and diagnostic: staged validation is used to localize the dominant bottleneck in one fixed implementation, not merely to report another benchmark comparison.

\section{System and Protocol}
Table~\ref{tab:setup} summarizes the tested configuration. The pipeline has two stages. First, a hierarchical VQ-VAE-2 codec compresses GPT-2 token sequences from length 64 to a short top-level code sequence of length 16. Second, MDLM is trained to generate those top-level codes in an absorbing-state discrete diffusion process. At inference time, generated codes are decoded back to text through the trained codec.

\begin{table}[t]
\centering
\twolinecaption{System configuration.}{The study uses one concrete implementation and reports practical diagnostics rather than a new training algorithm.}
\label{tab:setup}
\small
\begin{tabular}{@{}ll@{}}
\toprule
Component & Configuration \\
\midrule
Dataset & TinyStories \\
Tokenizer & GPT-2 tokenizer \\
Input length & 64 tokens \\
Codec & Hierarchical VQ-VAE-2 \\
Temporal compression & 64 to 32 to 16 \\
Top-level code length & 16 \\
Codebook sizes & 512 (top), 512 (down) \\
Generator & MDLM, SUBS parameterization \\
Primary scorer & External GPT-2 perplexity \\
Geometry variant & Additional geometry-aware regularization \\
\bottomrule
\end{tabular}
\end{table}

\begin{figure*}[!t]
\centering
\includegraphics[width=\textwidth]{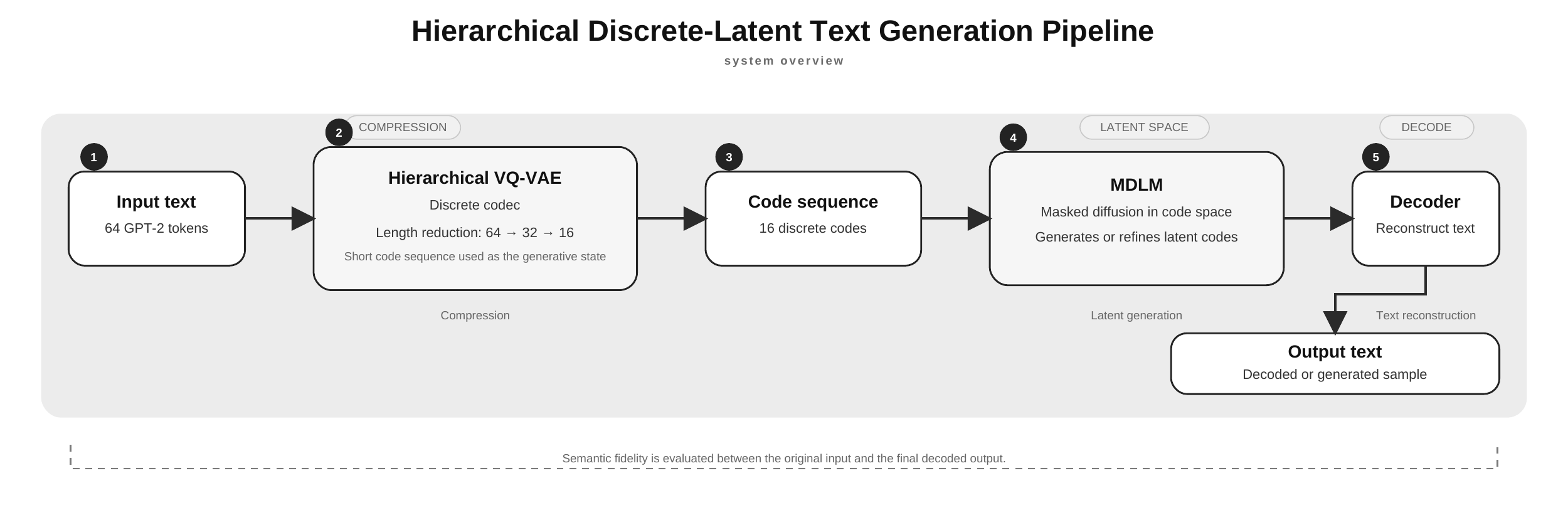}
\twolinecaption{Pipeline and readout order.}{Read the system left to right: first measure codec reconstruction loss, then compare generation modes under one scorer, and only then interpret auxiliary latent diagnostics.}
\label{fig:pipeline}
\end{figure*}

The evaluation is intentionally staged because different parts of the pipeline answer different engineering questions. First, paired codec reconstruction isolates what is lost before any latent generation occurs. Second, generator comparison under one shared scorer distinguishes representational loss from generation loss. Third, geometry-aware runs are interpreted only after the first two stages, because geometry is a secondary regularizer rather than the main system objective.

\begin{table}[t]
\centering
\twolinecaption{Artifact families.}{Each result in the paper is tied to one concrete set of released outputs rather than to an informal observation.}
\label{tab:artifact_flow}
\small
\begin{tabular}{@{}p{0.23\linewidth}p{0.22\linewidth}p{0.42\linewidth}@{}}
\toprule
Artifact family & Typical sample count & Main question answered \\
\midrule
Paired codec reconstructions & 256 paired samples & How much semantic fidelity is lost before latent generation? \\
Generation logs & 251--256 samples per mode & How do AR, token-space MDLM, and code-space MDLM compare under one scorer? \\
Geometry reports & 4 paired settings & Do latent proxy improvements transfer to decoded text quality? \\
\bottomrule
\end{tabular}
\end{table}

The external GPT-2 score is used as an anchor metric for stage-consistent comparison, not as a universal semantic metric. Its main advantage is that it can be applied to originals, reconstructions, and generated outputs under one fixed protocol. We therefore report both central tendency and tail statistics, while treating codebook activity and usage as health checks only. For the geometry study, we explicitly separate latent proxies from end-text metrics (SBERT, BERTScore, MAUVE, and the LLM judge) so that changes in local latent structure are not mistaken for improvements in decoded text quality.

A paired protocol is especially important in discrete-latent systems because averages can hide brittle failures. Pairing each source text with its codec reconstruction makes rare failures visible and allows a cleaner interpretation of where degradation starts. In practice, this also makes iteration cheaper: when a new codec checkpoint is trained, the paired reconstruction table can be recomputed before spending additional compute on the latent generator.

\section{Results}
\subsection{Codec bottleneck}
The codec is the first place where semantic fidelity can fail. Table~\ref{tab:recon_ppl} shows that reconstruction alone already causes a large shift in external perplexity. On the paired evaluation set, the mean rises from 16.24 for the original texts to 37.26 after codec reconstruction. The median rises from 15.17 to 27.36, which is an 80.4\% increase, and the reconstruction tail is much heavier ($p95$ 98.91 versus 25.10, a 294.1\% increase; max 105.59 versus 35.65).

\begin{table}[t]
\centering
\twolinecaption{Codec reconstruction and external perplexity.}{Even before latent generation, reconstruction alone introduces a large quality shift under the fixed scorer.}
\label{tab:recon_ppl}
\small
\begin{tabular}{@{}lrrrr@{}}
\toprule
Texts & Mean & Median & $p95$ & Max \\
\midrule
Original texts & 16.24 & 15.17 & 25.10 & 35.65 \\
Codec reconstructions & 37.26 & 27.36 & 98.91 & 105.59 \\
\bottomrule
\end{tabular}
\end{table}

This is the central empirical result of the paper. The codec does not exhibit a trivial collapse mode, yet reconstruction quality degrades sharply. In practical terms, this means that healthy code usage is not enough: a well-used codebook can still encode a representation that is difficult to decode back into semantically faithful text.

\begin{figure*}[!t]
\centering
\includegraphics[width=\textwidth]{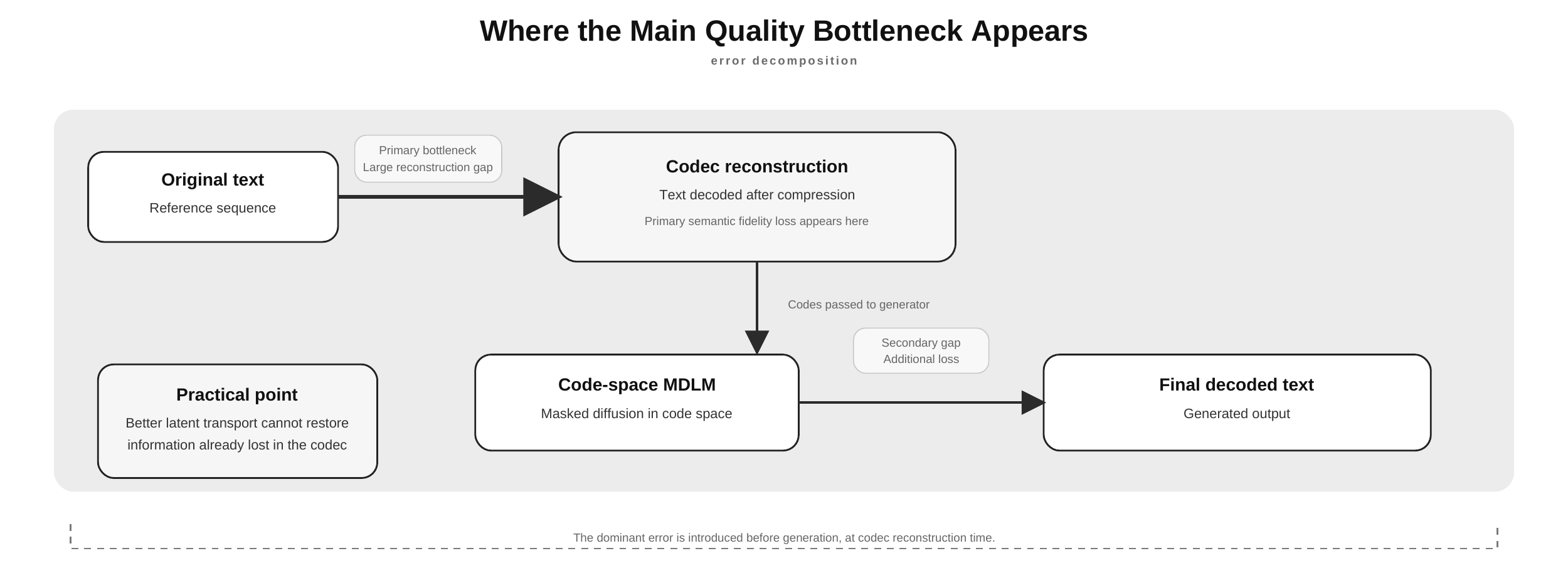}
\twolinecaption{Location of the main bottleneck.}{The dominant quality drop is already visible at codec reconstruction time; code-space generation adds additional loss but does not create the primary gap.}
\label{fig:bottleneck}
\end{figure*}

\subsection{Codebook diagnostics}
A natural first suspicion in discrete-latent systems is codebook collapse. The available artifacts do not support that explanation here. Both codec codebooks remain fully active, and the empirical code distribution is broad both during training and during generation.

\begin{table}[t]
\centering
\twolinecaption{Codebook and marginal diagnostics.}{The codebooks are well utilized, but these healthy statistics do not prevent large reconstruction degradation.}
\label{tab:codebook_health}
\small
\begin{tabular}{@{}lrrr@{}}
\toprule
Diagnostic & Top & Down & Generated top \\
\midrule
Active codes & 512 & 512 & 508 \\
Support size & 512 & 512 & 512 \\
Usage perplexity & 478.21 & 472.51 & 414.43 \\
$D_{\mathrm{KL}}(p_{\mathrm{train}}\|p_{\mathrm{gen}})$ & -- & -- & 0.0175 \\
\bottomrule
\end{tabular}
\end{table}

These statistics are therefore best read as coverage and utilization diagnostics. They can rule out trivial collapse, but they do not establish semantic fidelity. The large gap in Table~\ref{tab:recon_ppl} shows that these are different properties.

\subsection{Code-space vs. token-space diffusion}
Table~\ref{tab:external_ppl} compares three generation modes under the same external scoring protocol. The autoregressive reference remains best (median 23.27). Token-space MDLM is worst (median 38.42). The code-space pipeline is intermediate (median 26.55) and clearly improves over token-space diffusion: relative to token-space MDLM, it reduces median external perplexity by 30.9\%, lowers the mean by 32.9\%, and lowers $p95$ by 36.6\%.

\begin{table}[t]
\centering
\twolinecaption{External-perplexity comparison.}{Code-space diffusion improves substantially over token-space diffusion, but still trails the autoregressive reference.}
\label{tab:external_ppl}
\small
\begin{tabular}{@{}lrrrr@{}}
\toprule
Model & Count & Mean & Median & $p95$ \\
\midrule
AR baseline & 251 & 30.98 & 23.27 & 56.11 \\
MDLM only (token space) & 256 & 44.74 & 38.42 & 93.60 \\
MDLM plus codec (code space) & 256 & 30.01 & 26.55 & 59.36 \\
\bottomrule
\end{tabular}
\end{table}

This comparison supports two conclusions. First, generating in code space is not a pointless complication: it provides a substantial quality gain over token-space diffusion in the tested setup. Second, it supports the codec-limited interpretation of the whole system. Once the code-space generator is better than token-space MDLM but still below the autoregressive reference, the remaining gap is more plausibly attributed to the representational bottleneck than to a complete failure of the latent generator.

\subsection{Gap decomposition}
A useful way to read the same results is to compare every stage against the same reference point: the original texts scored by the external GPT-2 evaluator. Table~\ref{tab:gap_decomp} makes that comparison explicit using the median external perplexity.

\begin{table}[t]
\centering
\twolinecaption{Median-gap decomposition.}{Using one common reference point makes it easy to see where most quality loss enters the pipeline.}
\label{tab:gap_decomp}
\footnotesize
\setlength{\tabcolsep}{3.5pt}
\resizebox{\columnwidth}{!}{%
\begin{tabular}{@{}lrrr@{}}
\toprule
Evaluation point & Median ext. PPL & $\Delta$ vs. original & Rel. change \\
\midrule
Original texts & 15.17 & 0.00 & 0\% \\
Codec reconstructions & 27.36 & +12.19 & +80.4\% \\
Code-space MDLM output & 26.55 & +11.38 & +75.0\% \\
Token-space MDLM output & 38.42 & +23.25 & +153.3\% \\
\bottomrule
\end{tabular}
}
\end{table}

The decomposition does not prove a causal law, but it provides a clear engineering picture: most of the visible quality drop is already introduced before the latent generator has a chance to help. Code-space MDLM remains much closer to codec reconstruction than token-space diffusion does, which suggests that the denoiser is extracting most of the value available from the compressed representation.

\subsection{Geometry-aware training}
We also evaluated an exploratory geometry-regularized codec variant. The additional regularizer is intended to make local latent errors stay closer to target codes under a cosine-based neighborhood proxy. We compare baseline and geometry-regularized runs in four matched evaluations: code-level summaries for the 16-code and 8-code settings, and decoded-text summaries for the same two settings, using the same downstream decoder and evaluation scripts. This experiment is descriptive rather than inferential and is included to test whether cleaner latent neighborhoods translate into better decoded text.

\begin{table}[t]
\centering
\twolinecaption{Code-level geometry summary.}{Geometry regularization improves local latent proxies, but the corresponding text metrics do not improve in the current runs.}
\label{tab:geometry_training}
\footnotesize
\setlength{\tabcolsep}{3.2pt}
\resizebox{\columnwidth}{!}{%
\begin{tabular}{@{}lrrrrr@{}}
\toprule
Setting & $\Delta$SED & $\Delta$Geo@5 & $\Delta$SBERT & $\Delta$Judge & $\Delta$MAUVE \\
\midrule
16-code setting & -0.141 & +0.023 & -0.0056 & -0.335 & -0.0017 \\
8-code setting & -0.329 & +0.019 & -0.0332 & -0.325 & +0.0009 \\
\bottomrule
\end{tabular}%
}
\end{table}

At the code level, semantic error distance decreases and geohit@5 improves, indicating cleaner local latent behavior. However, these improvements do not carry over to end-text quality in the available runs: generation SBERT decreases, the LLM-judge score drops by about 0.33 points in both settings, and MAUVE changes are negligible.

\begin{table}[t]
\centering
\twolinecaption{Token-level geometry summary.}{Geometry-regularized training changes latent structure, but token-level text metrics remain flat or slightly worse in the tested runs.}
\label{tab:geometry_token}
\footnotesize
\setlength{\tabcolsep}{3.2pt}
\resizebox{\columnwidth}{!}{%
\begin{tabular}{@{}lrrrr@{}}
\toprule
Setting & $\Delta$BERT-F1 & $\Delta$SBERT & $\Delta$Judge & $\Delta$MAUVE \\
\midrule
16-code setting & -0.005 & -0.008 & -0.097 & -0.0002 \\
8-code setting & -0.008 & -0.012 & -0.118 & +0.0000 \\
\bottomrule
\end{tabular}%
}
\end{table}

\begin{figure*}[!t]
\centering
\includegraphics[width=\textwidth]{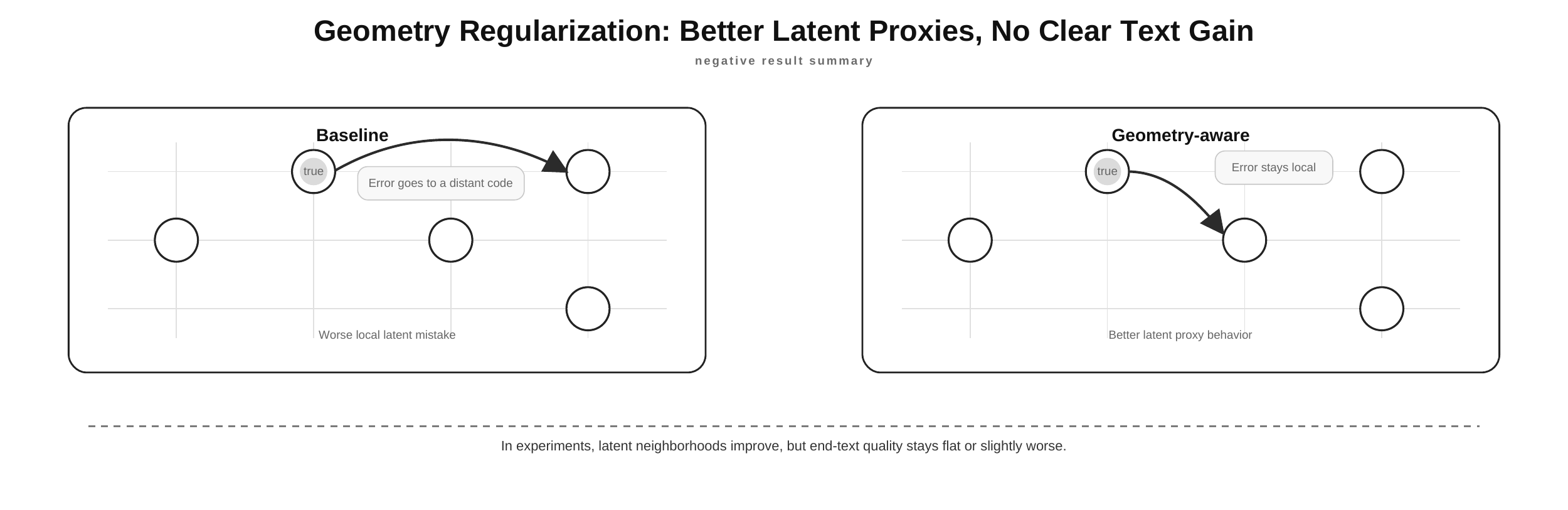}
\twolinecaption{Geometry-aware training as a negative result.}{The geometry-aware variant improves local latent proxy behavior, but the available runs do not show a corresponding gain in decoded text quality.}
\label{fig:geometry_negative}
\end{figure*}

The geometry results are therefore best read as a useful negative result. In the present implementation, the regularizer reshapes local latent structure but does not improve decoded text quality. For this reason, we treat geometry-aware training as a secondary representation-shaping tool rather than as a demonstrated quality-improvement mechanism.

\section{Discussion}
\subsection{Scope, metrics, and validity}
The study intentionally reports a controlled case study of one released implementation rather than a broad retraining campaign. The goal is bottleneck localization under fixed conditions, not leaderboard estimation across many settings. This scope is sufficient for diagnosing the dominant failure mode in the present artifacts, but it does not justify stronger claims about all hierarchical latent text generators.

Several limitations follow directly from that framing. The empirical evidence comes from one dataset (TinyStories), one compression regime (64-to-16), one codec family (VQ-VAE-2), and one generator type (MDLM). The reported comparisons are based on single runs and should therefore be read as descriptive evidence rather than as a multi-seed significance study. Likewise, the external GPT-2 scorer is used as a fixed shared diagnostic rather than as a universal semantic metric. Its value here is consistency across stages; it remains only one lens on quality, which is why we also report codebook diagnostics and, for the geometry comparison, additional text-level metrics.

\subsection{How to read the metrics}
A practical challenge in compressed text generation is that different metrics answer different questions, and they are easy to overinterpret. External perplexity is the main stage-consistent metric because it is available for originals, reconstructions, and generations under one scorer. Tail-aware variants of that metric matter because they expose brittle failures that are easy to miss in averages. Code usage and support size are health checks only: they can rule out trivial collapse, but they cannot establish semantic fidelity. Latent geometry proxies answer a different question again, namely whether local code-space structure changed.

\begin{table}[t]
\centering
\twolinecaption{Reading the main metric families.}{The table maps each metric family to the engineering question it actually answers.}
\label{tab:metric_roles}
\small
\begin{tabular}{@{}p{0.25\linewidth}p{0.27\linewidth}p{0.33\linewidth}@{}}
\toprule
Metric family & What it answers & What it does \emph{not} answer \\
\midrule
External PPL & End-text quality under one scorer & Whether the codebook is healthy \\
Tail PPL statistics & Reliability and brittle failures & Whether latent neighborhoods are cleaner \\
Usage / support size & Collapse and coverage & Whether meaning is preserved \\
Latent geometry proxies & Local code-space structure & Whether decoded text improved \\
\bottomrule
\end{tabular}
\end{table}

This split is especially important for the geometry study. If semantic error distance improves while SBERT, BERTScore, MAUVE, and the LLM judge do not, then the correct interpretation is that the representation changed without yielding a downstream text gain. In the current results, that is exactly what we observe.

\subsection{Why tail behavior matters}
The reconstruction gap is not only a shift in the mean; it is also a change in reliability. In the available paired outputs, most reconstructed samples remain interpretable, but a minority become severely degraded under the external scorer. This is why the analysis emphasizes the upper tail instead of reporting only one average number. In deployment-oriented short-text generation, occasional but extreme failures are often more costly than a modest average loss because they dominate user-visible trust and manual filtering overhead.

The practical interpretation is straightforward. A compressed latent generator can be attractive when its average score is acceptable, but it becomes much less attractive when the codec introduces brittle edge cases that the latent generator cannot recover from. The reconstruction table suggests exactly this pattern: the median worsens materially, and the upper tail worsens much more. For future iterations, any claimed improvement should therefore be checked not only for average reconstruction quality but also for whether it narrows the upper tail.

\subsection{Practical implications}
The current evidence supports a simple engineering roadmap. First, the codec should be improved, because it is the dominant bottleneck and every downstream experiment depends on it. Second, once reconstruction improves, code-space diffusion remains the most promising non-autoregressive path because it already outperforms token-space diffusion in the tested setup. Third, auxiliary regularizers such as geometry-aware losses should be prioritized only when they improve both their intended latent proxy and at least one paired decoded-text metric.

\begin{table}[t]
\centering
\twolinecaption{Failure taxonomy for staged debugging.}{The same system can fail in several qualitatively different ways; the table links each failure class to the most informative diagnostic.}
\label{tab:failure_taxonomy}
\small
\begin{tabular}{@{}p{0.18\linewidth}p{0.28\linewidth}p{0.18\linewidth}p{0.22\linewidth}@{}}
\toprule
Class & Typical symptom & Severity & Best first diagnostic \\
\midrule
Near-faithful paraphrase & Wording changes, core event retained & Low & Pairwise semantic similarity \\
Detail loss & Topic retained, key relation omitted & Medium & Paired reconstruction review \\
Semantic drift & Fluent output, meaning noticeably shifted & High & External PPL plus semantic judge \\
Catastrophic tail failure & Poor fluency or broken story logic & Very high & Tail quantiles and manual spot check \\
\bottomrule
\end{tabular}
\end{table}

This sample-level view explains why paired reconstruction is the first diagnostic to check after retraining. If the codec moves samples from near-faithful paraphrase into semantic drift or catastrophic tail failure, no downstream generator improvement can fully compensate. In that sense, the taxonomy simply restates the main empirical conclusion in a developer-friendly form: the codec sets the ceiling for the rest of the pipeline.

\subsection{Reproducibility and staged reporting}
The study is based on repository artifacts rather than a fresh retraining campaign. This is a deliberate design choice: it keeps the analysis auditable. The main evidence comes from released checkpoints, paired reconstruction outputs, generation logs, and summary tables derived from the same evaluation scripts, so the reported claims remain traceable to concrete files rather than to informal observations.

\begin{table}[t]
\centering
\twolinecaption{Artifact types used in the study.}{Each main claim is tied to one concrete family of files or logs.}
\label{tab:artifact_types}
\small
\begin{tabular}{@{}p{0.32\linewidth}p{0.53\linewidth}@{}}
\toprule
Claim & Primary artifact \\
\midrule
Codec bottleneck & Paired reconstruction PPL table \\
Codebook health & Code usage and marginal summaries \\
Generation comparison & AR and MDLM generation logs \\
Geometry summary & Baseline-vs-geometry reports \\
\bottomrule
\end{tabular}
\end{table}

In practice, this means the article can be reproduced in stages. A reader can first verify the paired codec reconstruction statistics, then check the generation comparison under the shared external scorer, and only after that inspect the geometry runs. This staged organization mirrors the logic of the paper itself: first isolate the codec, then compare generators, and only then interpret secondary regularization effects.

To keep future iterations disciplined, we summarize the staged promotion logic in one compact checklist. A new codec checkpoint should first improve paired reconstruction under the shared scorer without worsening the upper tail. Only then is it worth rerunning the downstream generator comparison. Likewise, a geometry loss should not be treated as successful until at least one paired decoded-text metric moves in the same favorable direction as the latent proxy.

\begin{table}[t]
\centering
\twolinecaption{Reusable reporting checklist.}{Each new model revision should be summarized in the same order so that promotion decisions remain comparable across runs.}
\label{tab:run_template}
\small
\begin{tabular}{@{}p{0.20\linewidth}p{0.23\linewidth}p{0.21\linewidth}p{0.24\linewidth}@{}}
\toprule
Stage & Must report & Gate question & Promote when \\
\midrule
Codec & Paired ext. PPL, median, $p95$ & Did reconstruction improve? & Both central tendency and tail improve or remain stable \\
Generator & AR vs token vs code-space ext. PPL & Is code-space still the best non-AR path? & Code-space remains ahead of token-space and does not widen the tail sharply \\
Diagnostics & Usage, support, $D_{\mathrm{KL}}$, latent proxies & Why did quality change? & Diagnostics explain text-level movement without replacing it \\
\bottomrule
\end{tabular}
\end{table}

\subsection{Resource and deployment perspective}
Although the present system is still far from a drop-in replacement for strong autoregressive generation, the architecture remains attractive for constrained environments. A shorter latent sequence can reduce the per-step categorical problem and may simplify batching or non-autoregressive refinement relative to direct token-space diffusion. The current results therefore support a nuanced conclusion: code-space generation is already preferable to token-space diffusion in the tested setup, but its remaining quality gap is conditioned on codec fidelity.

This component-level view matters for project planning. In many R\&D settings, the easiest changes are not the most useful ones. It is often simpler to add another regularizer or another latent metric than to revisit the codec and its reconstruction objective. The present evidence argues against that temptation. The shortest route to a stronger system is to improve the codec first, keep the code-space generator as the preferred non-autoregressive path, and treat auxiliary latent regularizers as optional until they demonstrate a direct text-level gain.

\begin{table}[t]
\centering
\twolinecaption{Implementation priorities.}{The table summarizes where additional engineering effort is likely to pay off first.}
\label{tab:engineering_priority}
\small
\begin{tabular}{@{}p{0.23\linewidth}p{0.23\linewidth}p{0.32\linewidth}@{}}
\toprule
Component & Current status & Highest-value next step \\
\midrule
Codec & Main bottleneck & Improve semantic reconstruction before changing the generator \\
Token-space MDLM & Weak reference & Use mainly as a lower baseline, not as the main path \\
Code-space MDLM & Operationally useful & Retain and re-evaluate after codec improvements \\
Geometry regularizer & Secondary, mixed outcome & Keep only if downstream text metrics begin to move \\
Evaluation pipeline & Already informative & Preserve the staged protocol for future runs \\
\bottomrule
\end{tabular}
\end{table}

\subsection{Practical use and follow-up}
The current system is not best viewed as a universal text generator. A more realistic interpretation is as a compressed generation stack for settings where short inputs, constrained compute, or fast non-autoregressive iteration matter. In such settings, the main operational question is not whether the latent pipeline matches the strongest autoregressive model on every metric, but whether it provides an acceptable quality-speed trade-off while remaining debuggable. The present results suggest that this is already true relative to token-space diffusion, but only when the codec is treated as a first-class production component.

Two immediate uses follow. For deployment-oriented short-text generation, code-space MDLM remains relevant because it improves materially over token-space diffusion while operating on a shorter sequence; however, paired reconstruction must act as a promotion gate, and a codec checkpoint that fails that gate should not be deployed even when codebook statistics look healthy. For research iteration, the same staged protocol shortens the loop: a run that fails reconstruction can be rejected before generator retraining, and a run that improves latent proxies without improving downstream text can be classified as representation-shaping rather than deployment-improving.

\begin{table}[t]
\centering
\twolinecaption{Decision matrix for practical use.}{The table summarizes when the current pipeline is useful and when it should remain a research prototype.}
\label{tab:use_matrix}
\small
\begin{tabular}{@{}p{0.25\linewidth}p{0.23\linewidth}p{0.38\linewidth}@{}}
\toprule
Condition & Suitability & Recommended action \\
\midrule
Short text, moderate quality tolerance & Moderate & Use code-space path after paired reconstruction check \\
Short text, strict fidelity requirement & Low & Prefer autoregressive baseline \\
Research iteration on latent methods & High & Use staged protocol to localize failures early \\
Token-space diffusion baseline study & High & Retain mainly as a lower reference point \\
\bottomrule
\end{tabular}
\end{table}

The next experiments are deliberately narrow. The first is a codec-only training revision that targets semantic fidelity under the same paired reconstruction protocol. The second is a repeated code-space generation comparison after that codec update. The third is a more decoder-aware geometry objective, evaluated only if the codec gap narrows. This order keeps the iteration loop aligned with the current evidence and avoids spending effort on improvements that the existing bottleneck would hide.

A useful methodological extension would also be a small multi-seed check for the geometry runs. The present geometry evidence is already informative as descriptive evidence, but a limited seed sweep would clarify whether the slight text-level degradations are stable or mostly noise. Even a stable negative result would remain valuable because it would identify which regularizers are not yet paying off in a compressed text-generation pipeline.

\subsection{Experiment triage}
When multiple metrics are available, the main practical challenge is deciding which ones should trigger another expensive training run. The current evidence suggests a simple triage rule. A codec variant should be promoted to full downstream evaluation only if it improves paired reconstruction under the shared external scorer and does not worsen the upper tail. If those two conditions are not met, additional generator experiments are unlikely to reveal a meaningful end-text gain.

A second triage rule concerns the role of latent-space diagnostics. Codebook usage, support size, and code-marginal divergence should be checked in every run because they are cheap and can reveal trivial collapse. However, once these diagnostics are already healthy, they should not be treated as success metrics on their own. In the present system, all of them look reasonable while reconstruction fidelity remains poor. This means they function best as safety checks, not as decision criteria for model selection.

A third triage rule concerns auxiliary objectives such as geometry-aware regularization. These objectives are worth retaining only when they satisfy a two-stage condition: they must improve their intended latent proxy, and at least one paired downstream text metric must move in the same favorable direction. If the first condition holds but the second does not, the regularizer may still be scientifically interesting, but it should not take priority over codec-focused work. This rule keeps the experimentation budget aligned with downstream utility.

In applied settings, these triage rules reduce the risk of pursuing attractive but low-value changes. They also make team communication easier. A short report can say, for example, that a run passed the codebook-health gate but failed the reconstruction gate, so no generator retraining was performed. That kind of disciplined workflow is often more useful than another round of broad hyperparameter tuning, especially when compute budgets and project timelines are tight.

\subsection{Inference path and stage attribution}
The pipeline can also be understood as a sequence of four transformations. First, a 64-token TinyStories sequence is compressed by the hierarchical codec into a 16-code latent. Second, MDLM predicts or refines that 16-code sequence in latent space. Third, the codec decoder expands those codes back into a 64-token text. Fourth, the external scorer evaluates the resulting text under the same procedure used for all systems in the paper. Each stage can fail for a different reason, and the current evidence shows that the earliest stage already introduces the largest quality loss.

This end-to-end view clarifies why the analysis emphasizes staging. If the codec has already discarded distinctions that matter for semantic fidelity, the latent generator is operating on a compromised representation. In that situation, even a stronger latent generator can only improve the part of the error budget that appears after compression. This is why the code-space path can still outperform token-space diffusion while remaining below the autoregressive reference: the generator contributes a real gain, but it cannot recover information that is no longer present in the latent.

\subsection{Why the latent path still matters}
One practical reason to study compressed latent generation even in the presence of an autoregressive quality gap is that the systems trade-off is still attractive. The code-space path reduces the modeled sequence from 64 tokens to 16 discrete symbols. This does not automatically make the system better, but it does change the optimization target from a longer token sequence to a much shorter latent one. In a production-oriented setting, that shorter sequence can simplify batching, reduce per-step categorical complexity, and make iterative denoising more manageable than direct token-space diffusion.

The present results therefore support a nuanced engineering conclusion. A compressed pipeline is already preferable to token-space diffusion in the tested setup, so the latent path is not merely an academic curiosity. At the same time, the remaining gap to the autoregressive reference shows that the gains from shorter latent sequences are conditional on codec fidelity. Compression changes where the compute is spent, but if the codec discards too much semantic information, the operational advantage is offset by the cost of lower quality and more failure handling.

\section{Conclusion}
We presented a staged validation protocol for hierarchical discrete-latent short-text generation and used it to localize the dominant bottleneck in one controlled implementation. The central claim is methodological: staged validation separates representational loss from generation loss and turns that diagnosis into explicit engineering decisions.

In the tested 64-to-16 pipeline, semantic fidelity is limited primarily by codec reconstruction under aggressive compression, even when the codebook remains healthy and well utilized. At the same time, code-space diffusion remains meaningfully better than token-space diffusion in the same setup, which supports continued work on compressed latent generation. Geometry-aware regularization changes local latent-space proxies but does not improve decoded text quality in the available runs.

The bounded takeaway is therefore straightforward. For this implementation, the clearest next step is improving codec semantic fidelity rather than relying on code usage or latent geometry alone. More broadly, the staged workflow itself should transfer more readily than any single numeric threshold: first test reconstruction, then compare generation paths, and only after that interpret auxiliary diagnostics.

This framing also clarifies how to read future improvements. A better codebook histogram, a cleaner latent neighborhood, or a narrower marginal gap may all be useful signals, but they are not substitutes for paired downstream text evidence. In the same way, a stronger denoiser can only help within the information preserved by the codec. The practical value of staged validation is therefore not only explanatory but procedural: it makes it easier to decide which experiments deserve another training cycle and which ones should stop after the first reconstruction check.

\balance

\end{document}